\documentclass[conference]{IEEEtran}
\IEEEoverridecommandlockouts
\usepackage{cite}
\usepackage{amsmath,amssymb,amsfonts}
\usepackage{graphicx}
\usepackage{textcomp}
\usepackage{xcolor}
\usepackage{hyperref}
\usepackage{url}

\usepackage[utf8]{inputenc}
\usepackage{times}
\usepackage{latexsym}
\usepackage{tabu}
\usepackage{array}
\usepackage{mathtools}
\usepackage{subcaption}

\usepackage{algorithm}
\usepackage{algpseudocode}
\usepackage{caption}
\usepackage{float}
\usepackage{flafter}

\usepackage{multirow}
\newcolumntype{L}[1]{>{\raggedright\let\newline\\\arraybackslash\hspace{0pt}}m{#1}}

\newcommand\BibTeX{B{\sc ib}\TeX}
\def\BibTeX{{\rm B\kern-.05em{\sc i\kern-.025em b}\kern-.08em
    T\kern-.1667em\lower.7ex\hbox{E}\kern-.125emX}}
\begin{document}

\title{Hyperparameter optimization with REINFORCE and Transformers
}

\author{\IEEEauthorblockN{Chepuri Shri Krishna}
\IEEEauthorblockA{
\textit{Walmart Global Tech India}\\
Bengaluru, India \\
Chepurishri.Krishna@walmartlabs.com}
\and
\IEEEauthorblockN{Ashish Gupta}
\IEEEauthorblockA{
\textit{Walmart Global Tech India}\\
Bengaluru, India \\
Ashish.Gupta@walmartlabs.com}
\and
\IEEEauthorblockN{Swarnim Narayan}
\IEEEauthorblockA{
\textit{Walmart Global Tech India}\\
Bengaluru, India \\
Swarnim.Narayan@walmartlabs.com}
\and
\IEEEauthorblockN{Himanshu Rai}
\IEEEauthorblockA{
\textit{Walmart Global Tech India}\\
Bengaluru, India \\
Himanshu.Rai@walmartlabs.com}
\and
\IEEEauthorblockN{Diksha Manchanda}
\IEEEauthorblockA{
\textit{Walmart Global Tech India}\\
Bengaluru, India \\
Diksha.Manchanda@walmartlabs.com}
\and
}

\maketitle

\begin{abstract}
Reinforcement Learning has yielded promising results for Neural Architecture Search (NAS). In this paper, we demonstrate how its performance can be improved by using a simplified Transformer block to model the policy network. The simplified Transformer uses a 2-stream attention-based mechanism to model hyper-parameter dependencies while avoiding layer normalization and position encoding. We posit that this parsimonious design balances model complexity against expressiveness, making it suitable for discovering optimal architectures in high-dimensional search spaces with limited exploration budgets. We demonstrate how the algorithm’s performance can be further improved by a) using an actor-critic style algorithm instead of plain vanilla policy gradient and b) ensembling Transformer blocks with shared parameters, each block conditioned on a different auto-regressive factorization order. Our algorithm works well as both a NAS and generic hyper-parameter optimization (HPO) algorithm: it outperformed most algorithms on NAS-Bench-101~\cite{ying2019bench}, a public data-set for benchmarking NAS algorithms. In particular, it outperformed RL based methods that use alternate architectures to model the policy network, underlining the value of using attention-based networks in this setting. As a generic HPO algorithm, it outperformed Random Search in discovering more accurate multi-layer perceptron model architectures across 2 regression tasks. We have adhered to guidelines listed in Lindauer and Hutter~\cite{lindauer2019best} while designing experiments and reporting results.
\end{abstract}


\section{Introduction}

Hyper-parameter optimization (HPO) is a key component of the ML model development cycle. Recent work has shown that older deep learning architectures such as the LSTM\cite{hochreiter1997long} and original GAN\cite{goodfellow2014generative} outperformed contemporary architectures on benchmark tasks after being extensively tuned for regularization and other hyper-parameters\cite{melis2017state,lucic2018gans}. 

HPO can be framed as the optimization of an unknown, possibly stochastic, objective function mapping from the hyper-parameter search space to a real valued scalar, the ML model’s accuracy or any other performance metric on the validation data-set. The search-space can extend beyond algorithm or architecture specific elements to encompass the space of data pre-processing and data-augmentation techniques, feature selections, as well as choice of algorithms. This is sometimes referred to as the CASH (Combined Algorithm Search and Hyperparameter tuning) problem\cite{thornton2013auto}. 

Neural Architecture Search (NAS) is a special type of HPO problem where the focus is on algorithm driven design of neural network architecture components or cells\cite{elsken2018neural}. Here, the search space is usually discrete and of variable dimensionality. Deep learning architectures designed via NAS algorithms have surpassed their hand-crafted counterparts for tasks such as image recognition and language modeling\cite{luo2018neural,zoph2018learning}, underlining the practical importance of this field of research. 

We present a new HPO algorithm, REINFORCE with Masked Attention Auto-regressive Density Estimators (ReMAADE), that uses a Transformer like architecture \cite{vaswani2017attention} to specify the policy network and employs Policy Gradient to tune its parameters. ReMAADE works well as both a NAS and generic HPO algorithm. For NAS, it outperforms most NAS algorithms on NASBench-101 \cite{ying2019bench}. It is able to discover better multi-layer perceptron models for regression tasks relative to Random Search on the Boston Housing and Naval Propulsion data-sets.

Our contributions can be summarized as:
\begin{itemize}
\item We present a 2-stream attention based architecture for capturing dependencies between hyper-parameters. This architecture's parametric complexity is invariant to the dimensionality of the search-space, yet the architecture is expressive enough to capture long range dependencies. 
\item We present an actor-critic style algorithm for stabilizing training and improving performance. This approach is useful in low computational budget settings.
\item We also investigate the effect of ensembling models with shared parameters, each conditioned on a different auto-regressive factorization order. We posit that this approach can be useful in large computational budget settings for discovering optimal architectures.
\end{itemize}

\begin{table}[t!]
\centering
\resizebox{0.49\textwidth}{!}
{
\begin{tabular}{l|c|}
\hline
Keyword & Description\\ 
\hline
NAS & Neural Architecture Search \\
\hline
ReMAADE & REINFORCE and Masked Attention Auto-Regressive Density Estimators \\
\hline
BANANAS & Bayesian Optimization with Neural Architectures for Neural Architecture Search \\
\hline
GP & Gaussian Process \\
\hline
MADE & Masked Autoregressive Density Estimator \\
\hline
MAADE & Masked Attention Autoregressive Density Estimator\\
\hline
NADE & Neural Autoregressive Density \\
\hline
PPO & Proximal Policy Optimization \\
\hline
NASBOT & Neural Architecture Search with Bayesian Optimization and optimal Transport\\
\hline
MCTS & Monte Carlo Tree Search\\
\hline
\end{tabular}
}
\caption{List of abbreviations}
\label{tab:abbreviations}
\end{table}

\section{Related Work}
HPO algorithms can be categorized under Bayesian Optimization, evolutionary learning, and gradient based methods, with random search widely regarded as a competitive baseline~\cite{bergstra2012random,li2019random}.

In methods based on Bayesian optimization \cite{kandasamy2018neural,falkner2018bohb,bergstra2011algorithms,white2019bananas}, the function mapping from the hyper-parameter search space to the validation score is modelled as a Gaussian Process (GP)\cite{williams2006gaussian}. This permits easy computation of the posterior distribution of the validation error for a new architecture. The performance of the method hinges on designing an appropriate kernel function for the GP and an acquisition function for fetching the next set of architectures for evaluation. A limitation of these approaches is that performance degrades in high-dimensional search spaces, because larger samples are required to update the posterior distribution \cite{fusi2018probabilistic}.

In evolutionary learning models, inspired by genetic evolution, the architectures are modelled as gene strings. The search proceeds by mutating and combining strings so as to hone in on promising architectures\cite{real2017large,real2019regularized,xie2017genetic}. Gradient based methods specify the objective function as a parametric model and proceed to optimize it with respect to the hyper-parameters via gradient-descent\cite{maclaurin2015gradient,luo2018neural,liu2018progressive}.

Model free RL based techniques \cite{zoph2016neural,zoph2018learning,baker2016designing,pham2018efficient,wang2018alphax} specify a policy network that learns to output desirable architectures. Search proceeds by training the policy network using Q-learning or policy gradient. These techniques are flexible in that they can search over variable length architectures, and have shown very promising results for neural architecture search.

NAS was initially formulated as a Reinforcement Learning (RL) problem, where a policy network was trained to sample more efficient architectures\cite{zoph2016neural}. In \cite{zoph2018learning}, a cell-based search in a search space of 13 operations is performed to find \emph{reduction} and \emph{normal} cells. These cells are then stacked to form larger networks. In \cite{baker2016designing}, the authors use Q-learning to discover promising architectures. Using a similar approach, \cite{zhong2018practical} performs NAS by sampling blocks of operations instead of cells, which can then be stacked to form networks. 

DARTS\cite{liu2018darts} deploys gradient-based search based on a continuous relaxation of an otherwise non-differentiable search space to discover high quality architectures. \cite{zela2019understanding,chen2019progressive,cai2018proxylessnas,xu2019pc} improve on this by using regularization to bridge the gap between validation and test scores.

Evolutionary learning algorithms have been applied to NAS as well. \cite{real2019regularized, liu2017hierarchical} discover high quality architectures for image classification using evolutionary learning techniques.

\section{Preliminaries}
Let the search space comprise $N$ hyper-parameters, indexed by $i \in \{1,\ldots, N \}$. RL methods based on policy gradient \cite{zoph2016neural,zoph2018learning} specify a policy network, parametrized by $\theta$, to learn a desired probability distribution over values of the hyperparameters, $P(a_1,a_2,\ldots,a_N; \theta)$, where $a_i$ denotes the value of the $i^{th}$ hyper-parameter.

We set up the training regime such that the policy network learns to assign higher probabilities to those sequences of hyperparameter values (henceforth, referred to as strings) that yield a higher accuracy on the cross-validation dataset. Accordingly, we maximize 
\begin{equation}
J(\theta) = \mathbb{E}_{a_{1:N}  \sim P(a_{1:N};\theta)} (f(a_{1:N}) - b)  
\label{eqn:gradient}
\end{equation}

where $f: \mathcal{H} \rightarrow \mathbb{R}$  is an unknown function that maps strings from the hyperparameter search space to the ML model's accuracy on the validation dataset. $b$ is a baseline function to reduce the variance of the estimate of the gradient of (\ref{eqn:gradient}).

We optimize the objective via gradient ascent where the gradient can be estimated using Reinforce~\cite{williams1992simple}:
\begin{equation}
\nabla J(\theta) \approx \frac{1}{m} \sum_{k=1}^m \nabla_\theta \log P(a_{1:n}^k; \theta)[f(a_{1:n}^k) - b]
\label{eqn:objective_gradient}
\end{equation}
where $a_{1:n}^k \sim P(a_{1:N};\theta)$. $f(a_{1:N})-b$ is referred to as the advantage function, denoted by $A(a_{1:N})$.

The optimization procedure alternates between two steps until we exhaust the exploration budget:
\begin{enumerate}

	\item sample a batch of action strings based on the current state of the policy network and fetch the corresponding rewards from the environment

	\item update the policy network’s parameters using policy gradient
\end{enumerate}
The exploration budget can be quantified in units of computation such as number of GPU/TPU hours for training architectures or, alternately, as the number of times the policy network can query the environment to fetch the architecture’s score. In case of the latter, it is assumed that all architectures consume identical compute for getting trained.

\section{Autoregressive Models for Density Estimation}
The policy network can be set up as an auto-regressive model, an approach that has been successfully applied to language models~\cite{mikolov2010recurrent,kim2016character}, generative models of images~\cite{oord2016pixel,salimans2017pixelcnn++,chen2017pixelsnail} and speech~\cite{oord2016wavenet}.

\begin{equation}
    P(a_{1:N};\theta) = \prod_{i=1}^{N}P(a_i|a_{1:i-1};\theta)
\label{eqn:autoregressive}
\end{equation}

The choice of a parametric architecture for modelling terms in (\ref{eqn:autoregressive}) becomes crucial as it needs to balance expressiveness against model complexity. The former is important to learn dependencies between hyper-parameters over increasing string lengths, while the latter needs to be economized for discovering optimal strings within an exploration  budget. RNN based networks struggle to learn adequate context representations over longer sequence lengths because of the vanishing gradient/exploding gradient problem~\cite{pascanu2013difficulty}. On the other hand, the parametric complexity of masked multi-layer perceptron based models, such as~MADE~\cite{germain2015made} and~NADE~\cite{uria2016neural}, increases with string length $N$ (table~\ref{tab:param_complexity}).

\subsection{Masked Attention Autoregressive Density Estimators (MAADE)}
\begin{figure}[h!]
\includegraphics[width=0.5\textwidth]{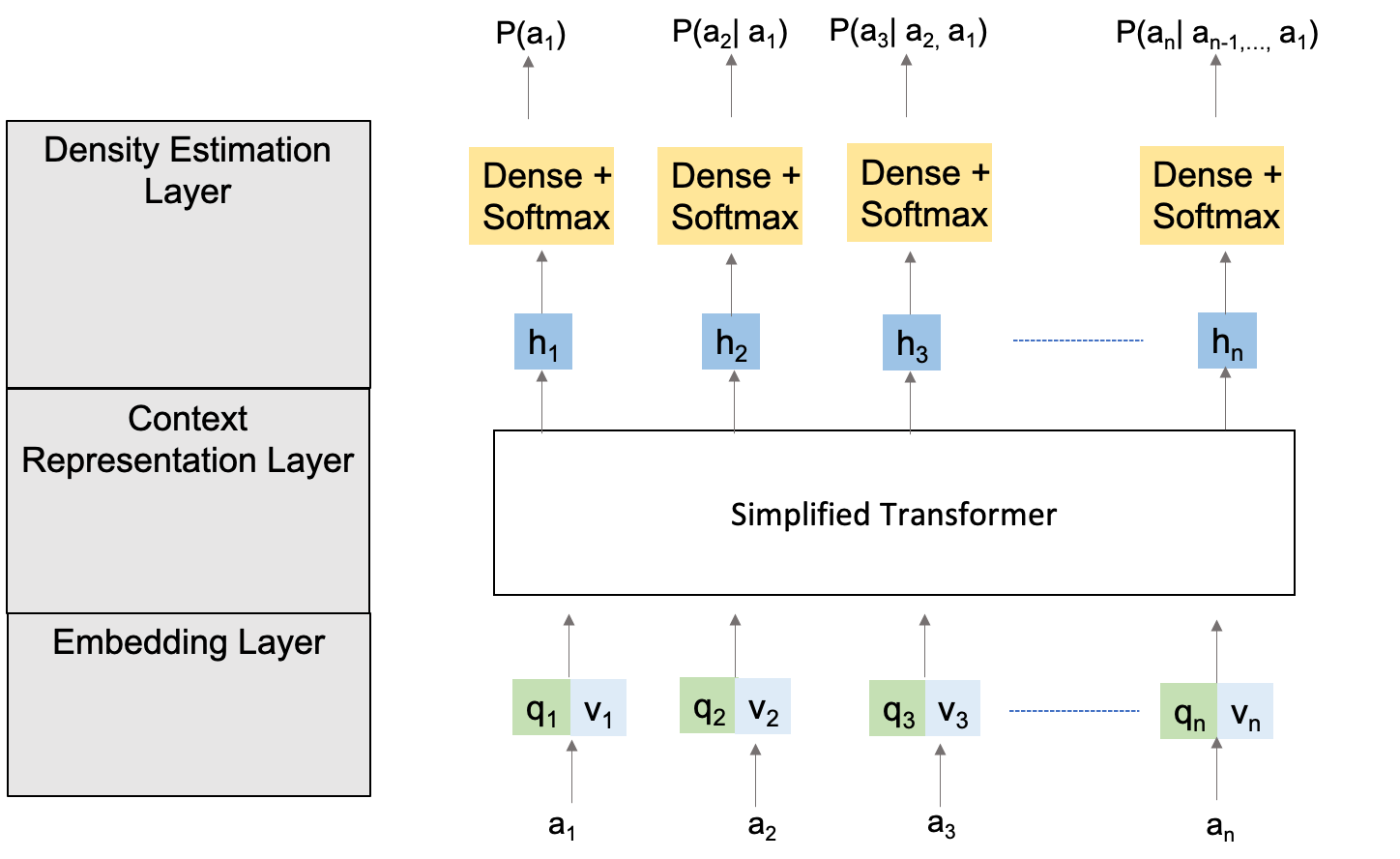}
\caption{Overall architecture layout}
\label{fig1}
\end{figure}

To strike a balance between expressiveness and complexity in designing the policy network, we propose the Masked Attention Autoregressive Density Estimation (MAADE) architecture, comprising three layers:
\begin{enumerate}
    \item \textbf{Embedding layer} maps hyper-parameters and hyper-parameter values to a $d$ dimensional vector space
\item \textbf{Context Representation layer} models dependencies between hyper-parameters as specified by the auto-regressive factorization
\item \textbf{Density Estimation Layer} computes the probability density for a string
\end{enumerate}

\subsubsection{\textbf{Embedding Layer}} Let $ q_i \in \mathbb{R}^d, \forall i=1, \cdots, N$, be query vectors for each of the $N$ hyper-parameters. We also maintain value vectors for the values that each hyper-parameter can take. We assume without loss of generality that all hyper-parameters assume categorical values in the same space and dimensionality $D$. We therefore share the embedding layer $V \in \mathbb{R}^{d \times D}$ across hyper-parameters. Given the value of the $i^{th}$ hyper-parameter, $a_{i}  \in \{1,\cdots,D \}$, the corresponding value vector is $V_{:,a_{i}} \in \mathbb{R}^d$  which we denote as $v_{i}$ with slight abuse of notation. Note that this framework can easily be extended to deal with hyper-parameters operating in different spaces as well, such that we maintain a separate embedding layer for each hyper-parameter family.

\subsubsection{\textbf{Context Representation Layer}} This layer produces $d$ dimensional contextual vectors, $h_i \in \mathbb{R}^d$ for each hyper-parameter as a function of the hyper-parameter being predicted and previously seen hyper-parameters:
\begin{equation}
    h_i \leftarrow H_\theta(q_i,q_{1:i-1},v_{1:i-1})
\end{equation}
Inspired by XLNet\cite{yang2019xlnet}, we use a two-stream masked attention based architecture comprising query and key vectors to compose $H_\theta$. A notable departure from XLNet is that since we are not predicting probabilities for a position but for a given hyper-parameter, we let the query vector of the target hyper-parameter attend to preceding key vectors. Each key vector attends to preceding key vectors as well as itself.

\begin{center}
\begin{figure}[h!]
\includegraphics[width=0.45\textwidth]{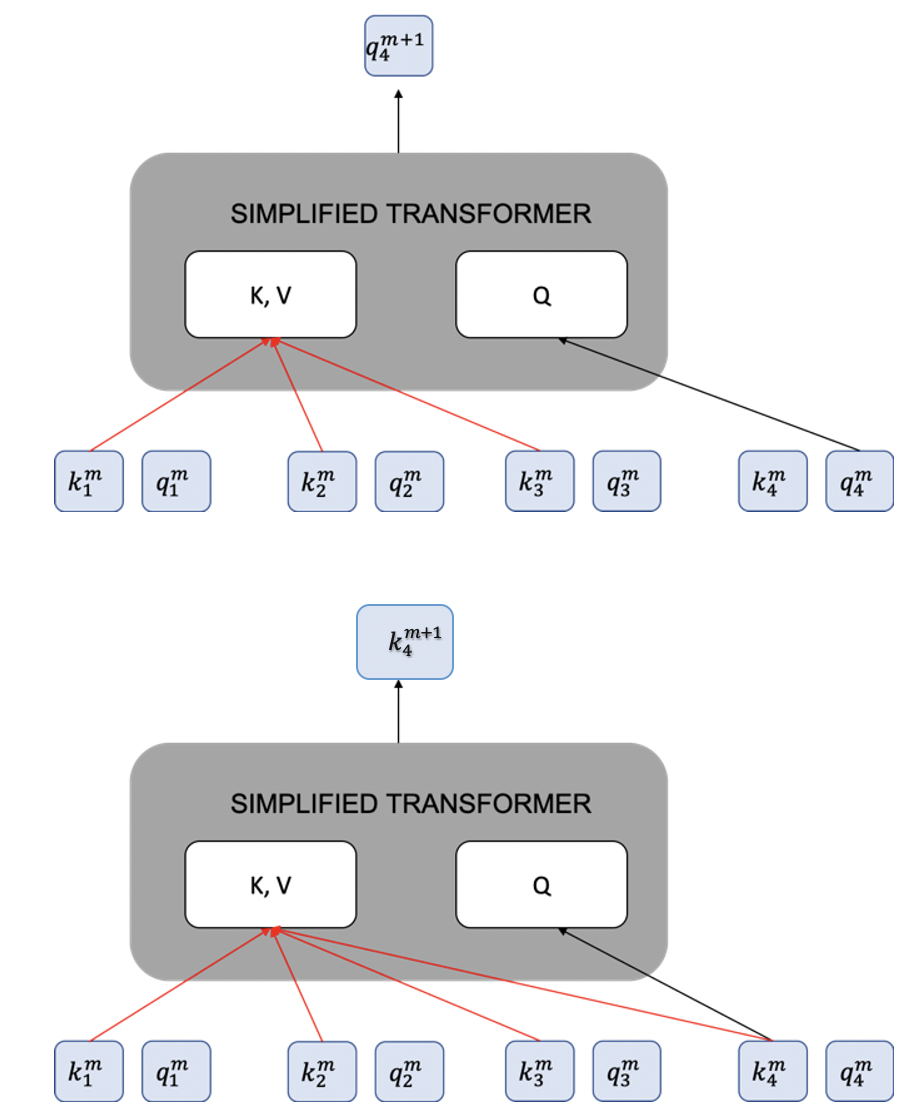}
\caption{2-stream attention}
\label{fig2}
\end{figure}
\end{center}

\begin{figure}[h!]
\centering
\includegraphics[height=2.4in,width=0.5\linewidth]{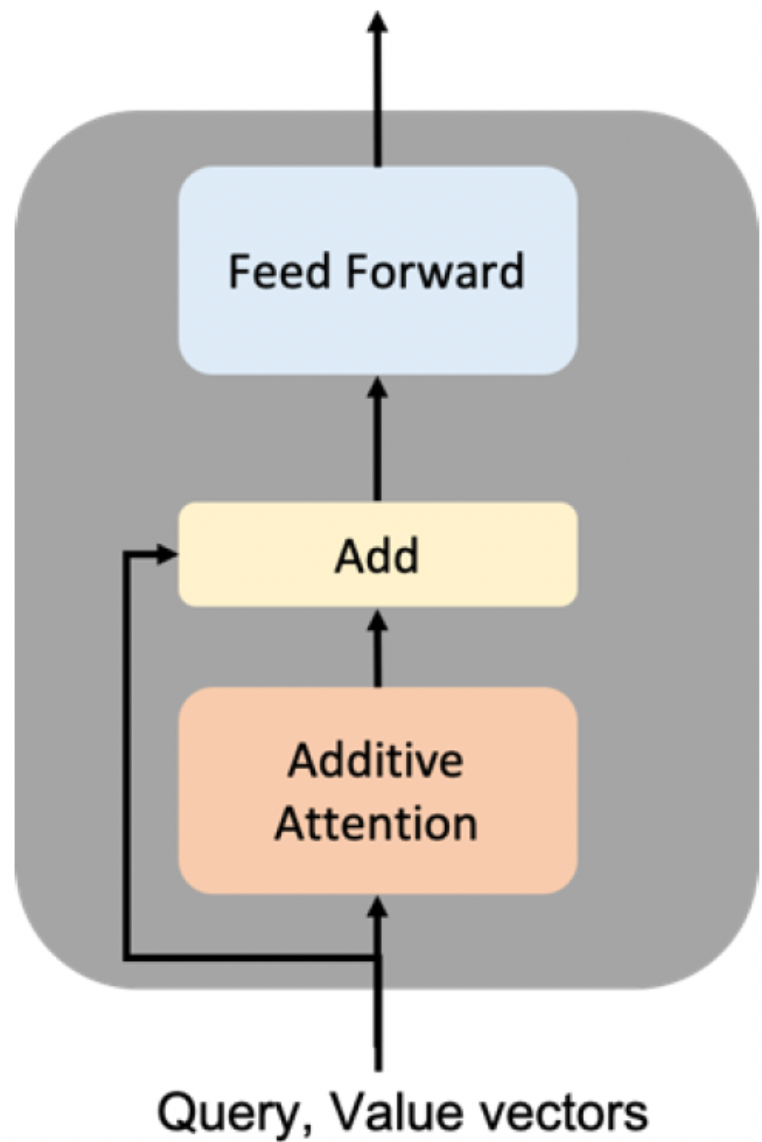}
\caption{Simplified Transformer block}
\label{fig:trans_block}
\end{figure}

Stream 1 is initialized as $q_i^{(0)} \leftarrow q_i$ stream 2 is initialized as $k_i^{(0)} \leftarrow v_i + q_i$. Then we update the streams as:
\begin{equation}
    q_i^{(m+1)} \leftarrow Tran(q_i^{m},k_{1:i-1}^{(m)})
\end{equation}
\begin{equation}
    k_i^{(m+1)} \leftarrow Tran(k_i^{m},k_{1:i}^{(m)})
\end{equation}
where \textit{Tran} is the transformer block referred in Figure \ref{fig:trans_block}. Finally, we get the contextual representations as: $h_i \leftarrow q_i^{(M)}$

To specify the Simplified Transformer block, we use a simplified version of Transformer~\cite{vaswani2017attention}. In the attention layer, we eschew dot production attention in favour of additive attention~\cite{bahdanau2014neural} to model interactions between query and key/value vectors as it was found to marginally improve the policy network's performance. We also found the policy network's performance to deteriorate when $M > 2$, obviating the need for both residual connections and layer-normalization. Finally, we do away with positional encoding since the sequence in which preceding hyper-parameters in the auto-regressive order were encountered doesn't matter.
\hfill

We provide details of the computation steps in the Simplified Transformer block:
\begin{multline}
    q_i^{(m+1)} \leftarrow PosFF(q_i^{(m)} + \\Masked\_Attention(Q=q_i^{(m)}, KV=k_{1:i-1}^{(m)}))\\  \forall i = 1,...,n
\end{multline}

\begin{multline}
    k_i^{(m+1)} \leftarrow PosFF(k_i^{(m)} + \\Masked\_Attention(Q=k_i^{(m)}, KV=k_{1:i}^{(m)})) \\ \forall i = 1,...,n
\end{multline}
\\
where \textit{Masked Attention} is the additive attention operation, and \textit{PosFF}, the position-wise feed-forward operation with \textit{relu} non-linearity replaced by \textit{tanh}:
\begin{equation}
    PosFF(x) = W_2(tanh(W_1x+b_1))+b_2
\end{equation}
Note that both the streams share parameters of the masked attention and feed-forward operations.

\vspace{3mm}
\subsubsection{\textbf{Density Estimation Layer}}:
In this layer, we pass the context representations through an affine transformation specific to the target hyper-parameter followed by a softmax,
\begin{equation}
    P(a_i|a_{1:i-1};\theta) = \frac{exp(h_{a_i}^TW_{a_i}^i+b_{a_i}^i)}{\sum_j exp(h_j^TW_j^i+b_j^i)}
    \label{eqn:softmax}
\end{equation}

\noindent\textbf{Complexity}
A key advantage of attention-based networks is that model complexity of the context representation layer doesn’t change with string length $N$. This is crucial for the NAS problem where the policy network needs to discover the best architecture within a limited exploration budget.\\
Table \ref{tab:param_complexity} defines the parametric complexity of each architecture.

\begin{table}[ht]
\centering
\resizebox{0.5\textwidth}{!}
{
\begin{tabular}{ |l|c|}
\hline
\textbf{Architecture} & \textbf{Complexity of Context Representation layer} \\ 
\hline
MAADE & $O(d^2)$\\
\hline
NADE & $O(Nd^2)$\\
\hline
MADE & $O(N^2d^2)$\\
\hline
LSTM based controller & $O(d^2)$\\
\hline
\end{tabular}
}
\caption{Parametric Complexity}
\label{tab:param_complexity}
\end{table}

\section{Training}
For training the policy network, we can optimize either (\ref{eqn:gradient}) or a PPO\cite{schulman2017proximal} objective as follows:

\begin{multline}
    J'(\theta) = \mathbb{E}_{a_{1:N} \sim P(a_{1:N};\theta)} [A(a_{1:N})min(r(a_{1:N}; \theta, \theta'), \\
    clip(r(a_{1:N};\theta, \theta'),1-\epsilon,1+\epsilon))]
    \label{eqn:ppo_objective}
\end{multline}
where,
\begin{equation}
    r(a_{1:N};\theta, \theta') = \frac{P(a_{1:N};\theta)}{P(a_{1:N};\theta')}
\end{equation}
An unbiased estimate of the gradient of (\ref{eqn:ppo_objective}) is:
\begin{multline}
    \nabla J'(\theta) \approx \frac{1}{m} \sum_{k=1}^m \nabla_\theta[A(a_{1:n}^k) min(r(a_{1:n}^k;\theta, \theta'),\\ clip(r(a_{1:n}^k;\theta,\theta'),1-\epsilon,1+\epsilon)]
\end{multline}
where $a_{1:N}^k \sim P(a_{1:N};\theta)$. We update the policy network parameters via gradient ascent:
\begin{equation}
    \theta \leftarrow \theta + \alpha\nabla J'(\theta)
    \label{eqn:gradient2}
\end{equation}
$\epsilon$, $B$ constitute ReMAADE’s hyper-parameters, which, along with the learning rate, $\alpha$, can be tuned via cross-validation. or the term in (\ref{eqn:gradient}). We can optionally add an entropy term to the objective to encourage exploration if we have a large exploration budget ~\cite{mnih2016asynchronous}. 

\subsection{ReMAADE algorithm}
We now describe the REINFORCE with Masked Attention Auto-regressive Density Estimators (ReMAADE) algorithm.\\
\textbf{Inputs}: 
\begin{itemize}
    \item Search space of architectures : $\mathcal{H}$
    \item Environment function that maps an architecture from the search-space to the corresponding validation accuracy : $f: a_{1:N} \in \mathcal{H} \rightarrow \mathbb{R}$
    \item Exploration budget: $E$
    \item ReMAADE hyperparameters: $B, S, \alpha, \epsilon $ where $B$ is the batch-size, $S$ is the set size of auto-regressive orderings to sample from, $\alpha$ is the learning rate, $\epsilon$ is the PPO co-efficient.
    \item MAADE hyperparameters: $d, M$ where $d$ is the embedding dimension for the transformer block and $M$ is the number of transformer blocks stacked.
\end{itemize}
The algorithm is described as follows:
  

\begin{algorithm}[h]
   \caption{\textit{ReMAADE}}
\begin{algorithmic}[1]
\State  $e\leftarrow 0$
\State $f(a^*)\leftarrow -\infty$
\State Initialize policy network and $\theta$
\While {$e<E$}
\State Sample $B$ valid hyper-parameter strings, $\{a_{1:N}^1, \cdots, a_{1:N}^B \}$, using the policy network
\State Fetch corresponding rewards, $\{f(a_{1:N}^1), \cdots, f(a_{1:N}^B) \}$
\State $a^* \leftarrow argmax(f(a^*),\{f(a_{1:N}^1), \cdots,f(a_{1:N}^B) \})$
\State Update $\theta$ as $\theta \leftarrow \theta + \alpha \nabla J'(\theta)$
\State $e\leftarrow e+B$
\EndWhile
\State Return best architecture found $a^*$
\label{alg:remaade}
\end{algorithmic}
\end{algorithm}

\begin{table*}[h!]
\centering
\resizebox{\linewidth}{!}
{
\begin{tabular}{ |c|c|c|c|c|}
\hline
\textbf{Algorithm} & \textbf{Source} & \textbf{Test Error (in \%)} & \textbf{Std-Deviation (in \%)} \\
\hline
TPE & Bergstra et al.\cite{bergstra2011algorithms} & 6.43 & 0.16 \\
\hline
BOHB & Falkner et al.\cite{falkner2018bohb} & 6.40 & 0.12 \\
\hline
Random Search & Bergstra et al.\cite{bergstra2012random} & 6.36 & 0.12 \\
\hline
NASBOT & Kandasamy et al.\cite{kandasamy2018neural} & 6.35 & 0.10 \\ 
\hline
Alpha X & Wang et al.\cite{wang2018alphax} & 6.31 & 0.13 \\
\hline
Reg Evolution & Real et al.\cite{real2019regularized} & 6.20 & 0.13 \\
\hline
ReMAADE & Ours & 6.16 & 0.25 \\ 
\hline
ReACTS & Ours & 6.13 & 0.25 \\ 
\hline
BANANAS & White et al.\cite{white2019bananas} & 5.77 & 0.31 \\
\hline
\end{tabular}
}
\caption{Performance of different search algorithms on NASBench-101 for short term run}
\label{tab:short_term}
\end{table*}

\section{ReACTS: Reducing variance using a critic}
Equation (\ref{eqn:objective_gradient}) provides an unbiased but high variance estimate of the gradient. The variance can be reduced by recourse to actor-critic algorithms~\cite{konda2000actor, schulman2015high}. We adopt a similar procedure, as follows. Define the value function as:
\begin{equation}
    V_\theta(a_{\leq i}) = \mathbb{E}_{a_{i+1:N}\sim P(a_{i+1:N}|a_{\leq i};\theta)}[f(a_{1:N})]
\end{equation}

In other words, the value function is the expected reward if we sample actions given the first $i$ actions, $a_1,a_2,\ldots,a_i$.\\
We can use Monte-Carlo policy evaluation to obtain an unbiased estimate of the value function. However, that would require us to query the environment further. Instead we take recourse to a simulator to estimate the value function as:
\begin{equation}
    \widehat{V_\theta}(a_{\leq i}) = \frac{1}{L}\sum_{l=1}^L S_\phi(a_{\leq i}, a^l_{>i}|a_{\leq i})\approx V_\theta(a_{\leq i})
    \label{eqn:reacts_policy}
\end{equation}
where $a^l_{> i}|a_{\leq i} \sim P(a_{i+1:N}|a_{\leq i};\theta)$. $S_\phi$ is a simulator, such as the meta-network in BANANAS\cite{white2019bananas}, that predicts the validation set reward associated with an architecture.\\
Then, an unbiased, lower variance (relative to equation (\ref{eqn:objective_gradient})) estimate of $\nabla J(\theta)$ can be computed as:
\begin{equation}
    \nabla J(\theta) \approx \frac{1}{B}\sum_{k=1}^B \sum_{i=1}^N \nabla_\theta log P(a^k_i|a^k_{< i};\theta)[f(a^k_{1:N}) - \widehat{V_\theta}(a^k_{< i})]
    \label{eqn:derivative}
\end{equation}
where $a_{1:N}^k \sim P(a_{1:N};\theta)$.
Search for the optimal architecture proceeds in discrete steps. At a given step $t$, we sample $B$ valid architectures using the policy network and fetch the corresponding rewards by querying the environment. Let the assembled data-set be denoted by
\begin{equation}
B_{\theta^t} := \{(f(a^1_{1:N}),a^1_{1:N}; \theta^t) ,\ldots  (f(a^B_{1:N}),a^B_{1:N}; \theta^t)\}
\end{equation}

We update $\phi$ such that it can predict rewards for architectures drawn from the policy network at step $t$:
\begin{equation}
    \phi = argmax(\mathbb{E}_{a_{1:N}\sim P(a_{1:N}; \theta^t)}  [-loss(S_\phi,f(a_{1:N}))])
\end{equation}
To make use of samples accumulated from earlier states of the policy network,$\{B_{\theta^0},B_{\theta^1}, \ldots,B_{\theta^{t}}\}$ we need to adjust for co-variate shift\cite{bickel2009discriminative}. Accordingly, we update $\phi$ as:
\begin{multline}
    \phi = argmax(\mathbb{E}_{a_{1:N}\sim P(a_{1:N};\theta^{0:t})}[-loss(S_\phi;f(a_{1:N}))\\\frac{(t+1)P(a_{1:N};\theta^t)}{P(a_{1:N};\theta^{0:t})}])
    \label{eqn:phi_update}
\end{multline}
The policy network’s state is then updated using equation (\ref{eqn:derivative}).

\subsection{ReACTS Algorithm}
We now have the machinery to describe ReACTS.

\begin{algorithm}[H]
\caption{\textit{ReACTS}}
\begin{algorithmic}[1]
\State $f(a^*)\leftarrow -\infty$
\State Initialize policy network parameters, $\theta^0$
\For {$t = 0\ldots T$}
\State 	Sample B valid hyper-parameter strings using policy network, Fetch corresponding rewards to assemble $B_{\theta^t}$
\State $a^* \leftarrow argmax(f(a^*),\{f(a_{1:N}^1), \cdots,f(a_{1:N}^B) \})$
\State Compute $\phi^t$ using $\{B_{\theta^0},B_{\theta^1}, \ldots, B_{\theta^t}\}$, equation (\ref{eqn:phi_update})
\State $
\theta^{t+1}\leftarrow ProcedureII(B_{\theta^t},\phi^t)
$
\EndFor
\State Return best architecture found $a^*$
\end{algorithmic}
\end{algorithm}

\begin{algorithm}[H]
\caption{\textit{ProcedureII}}
\begin{algorithmic}[1]
\State \textbf{Inputs:}  $B_{\theta^t}$, $S(\phi^t)$, $P(a_{1:N};\theta^t)$
\For {$k = 1,\ldots, B$}
\For {$i = 1,\ldots, N$}
\State Compute $\widehat{V_\theta}(a_{\leq i}^k)$  as per equation (\ref{eqn:reacts_policy})
\EndFor
\EndFor
\State Compute $\nabla J(\theta^t)$ as per equation (\ref{eqn:derivative})
\State $\theta^{t+1}\leftarrow \theta^t + \alpha\nabla J(\theta^t)$
\State Return $\theta^{t+1}$
\end{algorithmic}
\end{algorithm}

The \textbf{ReACTS} algorithm should outperform naive policy gradient-based methods if we have recourse to a good simulator. The Actor-Critic formula in equation (\ref{eqn:derivative}) fully exploits the Markovian nature of the implicit MDP of the policy network. Further, the transition dynamics are deterministic, eliminating another source of variance. Therefore, we expect this way of computing the gradient to reduce the variance of the gradient estimate and stabilize the training regime.

\section{Results on NASBench-101}
\vspace{2mm}
\noindent\textbf{NASBench-101 search space}
The NASBench-101 dataset~\cite{yang2019xlnet} is a public architecture dataset to facilitate NAS research and compare NAS algorithms. The search space comprises the elements of small-feed forward structures called cells. These cells are assembled together in a predefined manner to form an overall convolutional neural network architecture that is trained on the CIFAR-10 dataset.

A cell comprises 7 nodes, of which the first node is the input node and the last node is the output. The remaining 5 nodes need to be assigned one of 3 operations: 1x1 convolution, 3x3 convolution, or 3x3 max pooling. The nodes then need to be connected to form a valid directed acyclic graph (DAG). The NAS algorithm therefore needs to specify the operations for each of the 5 nodes, and then specify the edges to form a valid DAG. To limit the search space, NASBench-101 imposes additional constraints: the total number of edges cannot exceed 9, and there needs to be a path from the input node to the output node. This results in 423K valid and unique ways to specify a cell. NASBench-101 has pre-computed the validation and test errors for all the neural network architectures that can be designed from these 423K cell configurations.

To benchmark ReMAADE on NASBench-101, we investigate short term performance 
(exploration budget of 150 architectures). We include random search\cite{li2019random}, which is regarded as a competitive baseline, regularized evolution~\cite{real2019regularized}, and AlphaX, an RL algorithm that uses MCTS~\cite{wang2018alphax}. We also compare with several algorithms based on Bayesian optimization with GP priors: BOHB~\cite{snoek2012practical}, tree-structured Parzen estimator (TPE)~\cite{bergstra2011algorithms}, BANANAS~\cite{white2019bananas}, and NASBOT~\cite{kandasamy2018neural}. For all NAS algorithms, during a trial, we track the best random validation error achieved after $t$ explorations and the corresponding random test error. We report metrics averaged over 500 trials for each NAS algorithm. For ReMAADE, in all experiments, we set $M = 1, \epsilon = 0.1$, and used ADAM~\cite{adam} for updating $\theta$.

\subsection*{\textbf{Short Term Performance}}
NAS algorithms need to discover good architectures within 150 explorations to be of practical use. In this setting, (table~\ref{tab:short_term}), ReMAADE outperforms all algorithms with the exception of BANANAS. For ReMAADE, we set $\alpha = 1e-2, d = 36, S= 1, B=30$.





\subsection{Ablation studies}
\begin{table}[hbt!]
\centering
\resizebox{0.5\textwidth}{!}
{
\begin{tabular}{ |l|c|}
\hline
\textbf{Algorithm} & \textbf{Test Error (in \%)} \\ 
\hline
Random Search & 6.36 +-0.12\\
\hline
Plain vanilla REINFORCE & 6.26 +- 0.22\\
\hline
REINFORCE with MADE & 6.25 +- 0.22\\
\hline
ReMAADE w/o PPO & 6.16 +- 0.25\\
\hline
ReACTS w/o PPO & 6.13 +- 0.25\\
\hline
\end{tabular}
}
\caption{Ablation Study}
\label{tab:ablation_studies}
\end{table}
We perform an ablation study to understand the importance of the autoregressive component and MAADE in designing the context representation layer. We can use REINFORCE without an autoregressive model, which we call plain vanilla REINFORCE. In other words, we assume that all actions are independent:
\begin{equation}
    P(a_1,a_2,....,a_N;\theta) = \prod_i P(a_i)
\end{equation}
This amounts to updating only the bias terms in (\ref{eqn:softmax}) using policy gradient, and yields a baseline test set error of 6.26\%. Interestingly, using MADE~\cite{germain2015made} to design the autoregressive model failed to improve upon plain vanilla REINFORCE, underscoring the importance of MAADE in capturing auto-regressive dependencies. Using ReACTS marginally improved performance relative to ReMAADE (table~\ref{tab:ablation_studies}).

\subsection{Effect of auto-regressive ordering ensembles}
Autoregressive density estimation models struggle with terms in (\ref{eqn:autoregressive}) with $i >> 1$ since they use a fixed capacity in the context representation layer. This can be mitigated to some extent by picking an autoregressive factorization order that exploits spatial-temporal dependencies using an appropriate architecture. For instance, in generative modelling of images, the raster scan ordering is preferred as it is able to capture spatial dependencies in the immediate neighborhood~\cite{oord2016pixel}.

In case of neural architecture search, however, it is not clear, \textit{a priori}, what autoregressive order to fix. Therefore, training an ensemble of models, each with a different autoregressive factorization order, with parameters shared across all models, can potentially improve performance as shown in ~\cite{uria2016neural,yang2019xlnet}. To do so, we explicitly condition the density on the autoregressive factorization order and share the policy network’s parameters across orders.

Accordingly, we set up the following framework: Let $Z_{N!}$ denote the set of all possible permutations of length $N$ index sequences. Let $Z_s \subseteq Z_{N!}$ with set size $S$. Let $z_{s,t}$ denote the $t^{th}$ element of a permutation $z_S \in Z_S$. We then define the joint probability over the autoregressive factorization order and action string, $ (z_s,a_{1:N})$, as:
\begin{equation}
    P(z_s,a_{1:N};\theta) = \frac{1}{S} \prod_{i=1}^N {P(a_{z_{s,i}}|a_{z_{s,1:i-1}};\theta)}
\end{equation}
To sample from this distribution, we sample a permutation uniformly at random from $Z_s$. We then fix the autoregressive factorization order based on the sampled permutation and sample the action string based on the MAADE architecture.

Accordingly, we maximize the following PPO objective:
\begin{multline}
    J'(\theta) = \mathbb{E}_{z_s, a_{1:N} \sim P(z_s, a_{1:N};\theta, \theta')} [A(a_{1:N}) \\
    min(r(z_s,a_{1:N}; \theta, \theta'),clip(r(z_s, a_{1:N};\theta, \theta'),1-\epsilon,1+\epsilon))]
    \label{eqn:objective2}
\end{multline}
where, 
\begin{equation}
    r(z_s,a_{1:N}; \theta, \theta') = \frac{P(z_s,a_{1:N};\theta)}{P(z_s,a_{1:N};\theta')}
\end{equation}

Training proceeds as per the ReMAADE algorithm. We expect that higher values of $S$ will lead to improved performance as we increase the exploration budget. Also note that in the limit when $S=N!$ , the training objective is identical to the bidirectional training objective in~\cite{yang2019xlnet}.
We empirically validated this and found performance to improve as the number of orderings was increased to 4 and then deteriorate before asymptotically improving again, when given an exploration budget of 3,200 architectures on NAS-Bench 101 (ref Table~\ref{tab:order_ensembeling}).

\begin{table}[hbt!]
\centering
\resizebox{0.3\textwidth}{!}
{
\begin{tabular}{ |l|c|}
\hline
\textbf{S} & \textbf{Test Error(in \%)} \\ [3pt]
\hline
1 & 5.95  +-  0.18\\[3pt]
\hline
2 & 5.94  +-  0.19\\[3pt]
\hline
4 & 5.91  +-  0.19\\[3pt]
\hline
6 & 5.92  +-  0.19\\[3pt]
\hline
8 & 5.93  +-  0.19\\[3pt]
\hline
16 & 5.92  +-  0.19\\[3pt]
\hline
256 & 5.92  +-  0.19\\[3pt]
\hline
\end{tabular}
}
\caption{Effects of autoregressive factorization order ensembling}
\label{tab:order_ensembeling}
\end{table}
    












\section{Results for Hyper-parameter Optimization}
We evaluate the performance of ReMAADE in optimizing the hyper-parameters of a Multi-Layer Perceptron (MLP). The MLP model is trained on two real-world datasets for a regression task.

\begin{enumerate}
    \item \textbf{Boston Housing}\cite{harrison2015boston}: This data-set consists of 506 samples, each sample made of 13 scaled input variables and a scalar regression output, the housing price.
    \item \textbf{Naval propulsion plants}\cite{coraddu2016machine}: The data-set consists of 11,934 samples, each sample made of 16 scaled input variables and a scalar regression output variable (turbine degradation coefficient).
\end{enumerate}

ReMAADE is bench-marked against Random Search and TPE\cite{bergstra2015hyperopt}. 
Each hyper-parameter optimization algorithm is given a budget of 100 model explorations per trial, and the Root Mean Square Error (RMSE) obtained on the test-set is reported at the end of the trial. The 3 HPO algorithms are evaluated on RMSE averaged over 100 trials.

\begin{figure}
\centerline{\includegraphics[width=\linewidth]{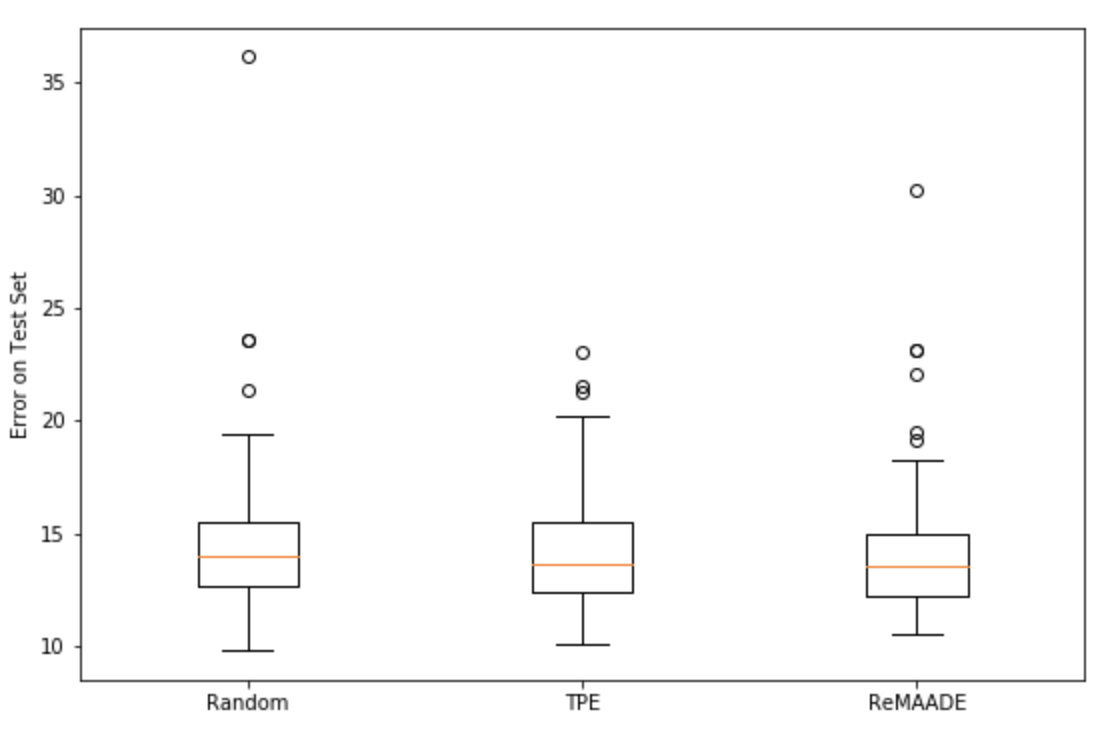}}
\caption{Benchmarking on the Boston Dataset}
\label{boxplotboston}
\end{figure}

\subsection{Case Study I: Boston Housing}
In this case-study, we train an MLP model on the Boston Housing data-set to predict house-prices. The MLP model has 10 hyper-parameters, including learning rate, l1 and l2 regularization, size of the hidden layer, number of iterations and choice of activation function. All 3 algorithms performed identically for this data-set and search space (Figure \ref{boxplotboston}).

\begin{figure}
\centerline{\includegraphics[width=\linewidth]{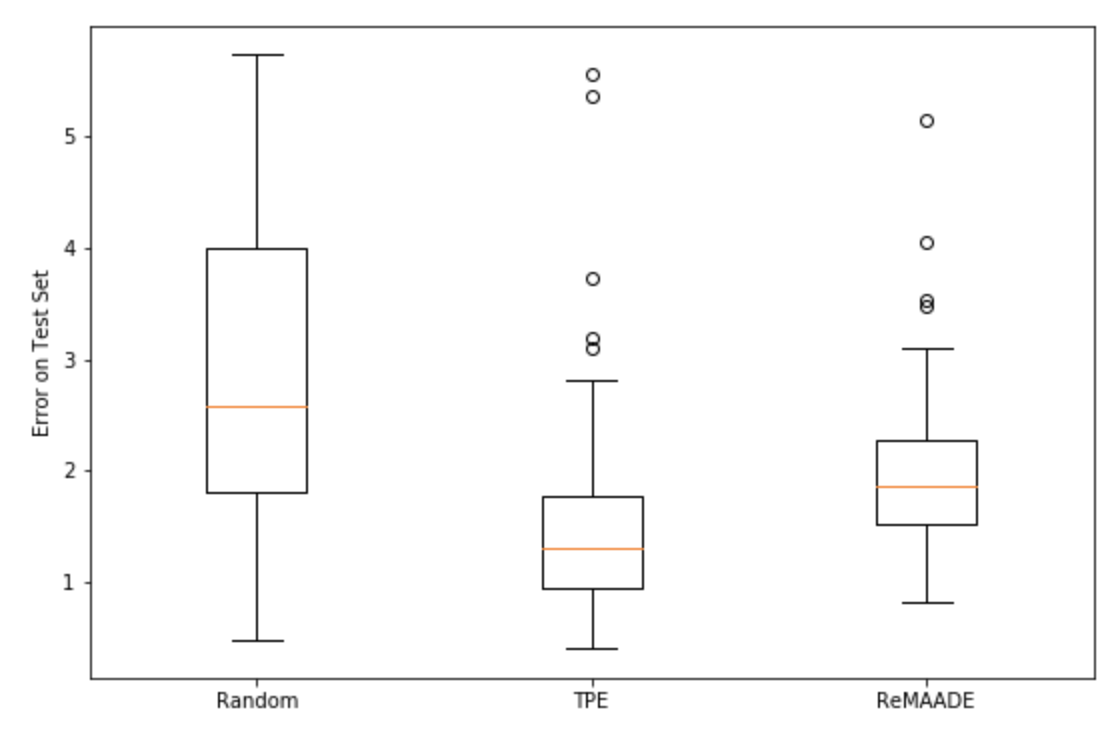}}
\caption{Benchmarking on the Naval Dataset}
\label{boxplotnaval}
\end{figure}

\subsection{Case Study II: Naval propulsion plants\cite{coraddu2016machine}}
In the second case study, we consider the task of improving condition based maintenance of naval propulsion plants. The data is generated using numerical simulator of a naval vessel which is characterized by a Gas Turbine (GT) propulsion plant. 
We trained a MLP model on this data-set with 10 hyper-parameters, including learning rate l1 and l2 penalties, size of hidden layer, number of iterations, activation function. In this setting, ReMAADE clearly outperforms Random Search and is competitive with TPE (Figure \ref{boxplotnaval}).




\subsection{Importance of Various Hyperparameters}

We find that Boston housing dataset with numerical features gave RMSE of about 14.102 while Naval Propulsion dataset gave 1.98 as RMSE with numerical features. Other search algorithms shown as baselines performed worse than ReMAADE, highlighting the use of Reinforced and masked attention auto-regressive ordering.

Among other algorithms which are shown in Table \ref{tab:short_term}, we compared our results with Random and TPE(Tree-structured Parzen Estimator) we got performance boost of 3.48\% in RMSE for Boston Housing dataset and nearly 10\% improvement in RMSE for Naval Propulsion dataset. Finally, we also observe that leveraging attention mechanism with auto-regressive ordering across different metadata helps the algorithm improve performance.

\section{Future Work}
We conclude that attention based auto-regressive models combined with policy gradient can be used as an effective hyper-parameter optimization problem. We need to further investigate the impact of conditioning on and ensembling across multiple autoregressive factorization orders on performance given large computational budgets. We also plan to investigate the performance of ReMAADE on other neural architecture search spaces such as DARTS~\cite{liu2018darts}. Another interesting line of work would be to use deep generative graph models with Policy Gradient\cite{you2018graph} to discover optimal NAS architectures. To model the graph context and capture node dependencies, we plan to investigate attention based mechanisms as was done in this paper.

\bibliographystyle{IEEEtran}
\bibliography{sample-base.bib}

\begin{thebibliography}{10}
\providecommand{\url}[1]{#1}
\csname url@samestyle\endcsname
\providecommand{\newblock}{\relax}
\providecommand{\bibinfo}[2]{#2}
\providecommand{\BIBentrySTDinterwordspacing}{\spaceskip=0pt\relax}
\providecommand{\BIBentryALTinterwordstretchfactor}{4}
\providecommand{\BIBentryALTinterwordspacing}{\spaceskip=\fontdimen2\font plus
\BIBentryALTinterwordstretchfactor\fontdimen3\font minus
  \fontdimen4\font\relax}
\providecommand{\BIBforeignlanguage}[2]{{%
\expandafter\ifx\csname l@#1\endcsname\relax
\typeout{** WARNING: IEEEtran.bst: No hyphenation pattern has been}%
\typeout{** loaded for the language `#1'. Using the pattern for}%
\typeout{** the default language instead.}%
\else
\language=\csname l@#1\endcsname
\fi
#2}}
\providecommand{\BIBdecl}{\relax}
\BIBdecl

\bibitem{ying2019bench}
C.~Ying, A.~Klein, E.~Real, E.~Christiansen, K.~Murphy, and F.~Hutter,
  ``Nas-bench-101: Towards reproducible neural architecture search,''
  \emph{arXiv preprint arXiv:1902.09635}, 2019.

\bibitem{lindauer2019best}
M.~Lindauer and F.~Hutter, ``Best practices for scientific research on neural
  architecture search,'' \emph{arXiv preprint arXiv:1909.02453}, 2019.

\bibitem{hochreiter1997long}
S.~Hochreiter and J.~Schmidhuber, ``Long short-term memory,'' \emph{Neural
  computation}, vol.~9, no.~8, pp. 1735--1780, 1997.

\bibitem{goodfellow2014generative}
I.~Goodfellow, J.~Pouget-Abadie, M.~Mirza, B.~Xu, D.~Warde-Farley, S.~Ozair,
  A.~Courville, and Y.~Bengio, ``Generative adversarial nets,'' in
  \emph{Advances in neural information processing systems}, 2014, pp.
  2672--2680.

\bibitem{melis2017state}
G.~Melis, C.~Dyer, and P.~Blunsom, ``On the state of the art of evaluation in
  neural language models,'' \emph{arXiv preprint arXiv:1707.05589}, 2017.

\bibitem{lucic2018gans}
M.~Lucic, K.~Kurach, M.~Michalski, S.~Gelly, and O.~Bousquet, ``Are gans
  created equal? a large-scale study,'' in \emph{Advances in neural information
  processing systems}, 2018, pp. 700--709.

\bibitem{thornton2013auto}
C.~Thornton, F.~Hutter, H.~H. Hoos, and K.~Leyton-Brown, ``Auto-weka: Combined
  selection and hyperparameter optimization of classification algorithms,'' in
  \emph{Proceedings of the 19th ACM SIGKDD international conference on
  Knowledge discovery and data mining}, 2013, pp. 847--855.

\bibitem{elsken2018neural}
T.~Elsken, J.~H. Metzen, and F.~Hutter, ``Neural architecture search: A
  survey,'' \emph{arXiv preprint arXiv:1808.05377}, 2018.

\bibitem{luo2018neural}
R.~Luo, F.~Tian, T.~Qin, E.~Chen, and T.-Y. Liu, ``Neural architecture
  optimization,'' in \emph{Advances in neural information processing systems},
  2018, pp. 7816--7827.

\bibitem{zoph2018learning}
B.~Zoph, V.~Vasudevan, J.~Shlens, and Q.~V. Le, ``Learning transferable
  architectures for scalable image recognition,'' in \emph{Proceedings of the
  IEEE conference on computer vision and pattern recognition}, 2018, pp.
  8697--8710.

\bibitem{vaswani2017attention}
A.~Vaswani, N.~Shazeer, N.~Parmar, J.~Uszkoreit, L.~Jones, A.~N. Gomez,
  {\L}.~Kaiser, and I.~Polosukhin, ``Attention is all you need,'' in
  \emph{Advances in neural information processing systems}, 2017, pp.
  5998--6008.

\bibitem{bergstra2012random}
J.~Bergstra and Y.~Bengio, ``Random search for hyper-parameter optimization,''
  \emph{Journal of machine learning research}, vol.~13, no. Feb, pp. 281--305,
  2012.

\bibitem{li2019random}
L.~Li and A.~Talwalkar, ``Random search and reproducibility for neural
  architecture search,'' \emph{arXiv preprint arXiv:1902.07638}, 2019.

\bibitem{kandasamy2018neural}
K.~Kandasamy, W.~Neiswanger, J.~Schneider, B.~Poczos, and E.~P. Xing, ``Neural
  architecture search with bayesian optimisation and optimal transport,'' in
  \emph{Advances in Neural Information Processing Systems}, 2018, pp.
  2016--2025.

\bibitem{falkner2018bohb}
S.~Falkner, A.~Klein, and F.~Hutter, ``Bohb: Robust and efficient
  hyperparameter optimization at scale,'' \emph{arXiv preprint
  arXiv:1807.01774}, 2018.

\bibitem{bergstra2011algorithms}
J.~S. Bergstra, R.~Bardenet, Y.~Bengio, and B.~K{\'e}gl, ``Algorithms for
  hyper-parameter optimization,'' in \emph{Advances in neural information
  processing systems}, 2011, pp. 2546--2554.

\bibitem{white2019bananas}
C.~White, W.~Neiswanger, and Y.~Savani, ``Bananas: Bayesian optimization with
  neural architectures for neural architecture search,'' \emph{arXiv preprint
  arXiv:1910.11858}, 2019.

\bibitem{williams2006gaussian}
C.~K. Williams and C.~E. Rasmussen, \emph{Gaussian processes for machine
  learning}.\hskip 1em plus 0.5em minus 0.4em\relax MIT press Cambridge, MA,
  2006, vol.~2, no.~3.

\bibitem{fusi2018probabilistic}
N.~Fusi, R.~Sheth, and M.~Elibol, ``Probabilistic matrix factorization for
  automated machine learning,'' in \emph{Advances in neural information
  processing systems}, 2018, pp. 3348--3357.

\bibitem{real2017large}
E.~Real, S.~Moore, A.~Selle, S.~Saxena, Y.~L. Suematsu, J.~Tan, Q.~V. Le, and
  A.~Kurakin, ``Large-scale evolution of image classifiers,'' in
  \emph{Proceedings of the 34th International Conference on Machine
  Learning-Volume 70}.\hskip 1em plus 0.5em minus 0.4em\relax JMLR. org, 2017,
  pp. 2902--2911.

\bibitem{real2019regularized}
E.~Real, A.~Aggarwal, Y.~Huang, and Q.~V. Le, ``Regularized evolution for image
  classifier architecture search,'' in \emph{Proceedings of the AAAI conference
  on artificial intelligence}, vol.~33, 2019, pp. 4780--4789.

\bibitem{xie2017genetic}
L.~Xie and A.~Yuille, ``Genetic cnn,'' in \emph{Proceedings of the IEEE
  international conference on computer vision}, 2017, pp. 1379--1388.

\bibitem{maclaurin2015gradient}
D.~Maclaurin, D.~Duvenaud, and R.~Adams, ``Gradient-based hyperparameter
  optimization through reversible learning,'' in \emph{International Conference
  on Machine Learning}, 2015, pp. 2113--2122.

\bibitem{liu2018progressive}
C.~Liu, B.~Zoph, M.~Neumann, J.~Shlens, W.~Hua, L.-J. Li, L.~Fei-Fei,
  A.~Yuille, J.~Huang, and K.~Murphy, ``Progressive neural architecture
  search,'' in \emph{Proceedings of the European Conference on Computer Vision
  (ECCV)}, 2018, pp. 19--34.

\bibitem{zoph2016neural}
B.~Zoph and Q.~V. Le, ``Neural architecture search with reinforcement
  learning,'' \emph{arXiv preprint arXiv:1611.01578}, 2016.

\bibitem{baker2016designing}
B.~Baker, O.~Gupta, N.~Naik, and R.~Raskar, ``Designing neural network
  architectures using reinforcement learning,'' \emph{arXiv preprint
  arXiv:1611.02167}, 2016.

\bibitem{pham2018efficient}
H.~Pham, M.~Y. Guan, B.~Zoph, Q.~V. Le, and J.~Dean, ``Efficient neural
  architecture search via parameter sharing,'' \emph{arXiv preprint
  arXiv:1802.03268}, 2018.

\bibitem{wang2018alphax}
L.~Wang, Y.~Zhao, Y.~Jinnai, and R.~Fonseca, ``Alphax: exploring neural
  architectures with deep neural networks and monte carlo tree search,''
  \emph{arXiv preprint arXiv:1805.07440}, 2018.

\bibitem{zhong2018practical}
Z.~Zhong, J.~Yan, W.~Wu, J.~Shao, and C.-L. Liu, ``Practical block-wise neural
  network architecture generation,'' in \emph{Proceedings of the IEEE
  conference on computer vision and pattern recognition}, 2018, pp. 2423--2432.

\bibitem{liu2018darts}
H.~Liu, K.~Simonyan, and Y.~Yang, ``Darts: Differentiable architecture
  search,'' \emph{arXiv preprint arXiv:1806.09055}, 2018.

\bibitem{zela2019understanding}
A.~Zela, T.~Elsken, T.~Saikia, Y.~Marrakchi, T.~Brox, and F.~Hutter,
  ``Understanding and robustifying differentiable architecture search,''
  \emph{arXiv preprint arXiv:1909.09656}, 2019.

\bibitem{chen2019progressive}
X.~Chen, L.~Xie, J.~Wu, and Q.~Tian, ``Progressive differentiable architecture
  search: Bridging the depth gap between search and evaluation,'' in
  \emph{Proceedings of the IEEE International Conference on Computer Vision},
  2019, pp. 1294--1303.

\bibitem{cai2018proxylessnas}
H.~Cai, L.~Zhu, and S.~Han, ``Proxylessnas: Direct neural architecture search
  on target task and hardware,'' \emph{arXiv preprint arXiv:1812.00332}, 2018.

\bibitem{xu2019pc}
Y.~Xu, L.~Xie, X.~Zhang, X.~Chen, G.-J. Qi, Q.~Tian, and H.~Xiong, ``Pc-darts:
  Partial channel connections for memory-efficient architecture search,'' in
  \emph{International Conference on Learning Representations}, 2019.

\bibitem{liu2017hierarchical}
H.~Liu, K.~Simonyan, O.~Vinyals, C.~Fernando, and K.~Kavukcuoglu,
  ``Hierarchical representations for efficient architecture search,''
  \emph{arXiv preprint arXiv:1711.00436}, 2017.

\bibitem{williams1992simple}
R.~J. Williams, ``Simple statistical gradient-following algorithms for
  connectionist reinforcement learning,'' \emph{Machine learning}, vol.~8, no.
  3-4, pp. 229--256, 1992.

\bibitem{mikolov2010recurrent}
T.~Mikolov, M.~Karafi{\'a}t, L.~Burget, J.~{\v{C}}ernock{\`y}, and
  S.~Khudanpur, ``Recurrent neural network based language model,'' in
  \emph{Eleventh annual conference of the international speech communication
  association}, 2010.

\bibitem{kim2016character}
Y.~Kim, Y.~Jernite, D.~Sontag, and A.~M. Rush, ``Character-aware neural
  language models,'' in \emph{Thirtieth AAAI Conference on Artificial
  Intelligence}, 2016.

\bibitem{oord2016pixel}
A.~v.~d. Oord, N.~Kalchbrenner, and K.~Kavukcuoglu, ``Pixel recurrent neural
  networks,'' \emph{arXiv preprint arXiv:1601.06759}, 2016.

\bibitem{salimans2017pixelcnn++}
T.~Salimans, A.~Karpathy, X.~Chen, and D.~P. Kingma, ``Pixelcnn++: Improving
  the pixelcnn with discretized logistic mixture likelihood and other
  modifications,'' \emph{arXiv preprint arXiv:1701.05517}, 2017.

\bibitem{chen2017pixelsnail}
X.~Chen, N.~Mishra, M.~Rohaninejad, and P.~Abbeel, ``Pixelsnail: An improved
  autoregressive generative model,'' \emph{arXiv preprint arXiv:1712.09763},
  2017.

\bibitem{oord2016wavenet}
A.~v.~d. Oord, S.~Dieleman, H.~Zen, K.~Simonyan, O.~Vinyals, A.~Graves,
  N.~Kalchbrenner, A.~Senior, and K.~Kavukcuoglu, ``Wavenet: A generative model
  for raw audio,'' \emph{arXiv preprint arXiv:1609.03499}, 2016.

\bibitem{pascanu2013difficulty}
R.~Pascanu, T.~Mikolov, and Y.~Bengio, ``On the difficulty of training
  recurrent neural networks,'' in \emph{International conference on machine
  learning}, 2013, pp. 1310--1318.

\bibitem{germain2015made}
M.~Germain, K.~Gregor, I.~Murray, and H.~Larochelle, ``Made: Masked autoencoder
  for distribution estimation,'' in \emph{International Conference on Machine
  Learning}, 2015, pp. 881--889.

\bibitem{uria2016neural}
B.~Uria, M.-A. C{\^o}t{\'e}, K.~Gregor, I.~Murray, and H.~Larochelle, ``Neural
  autoregressive distribution estimation,'' \emph{The Journal of Machine
  Learning Research}, vol.~17, no.~1, pp. 7184--7220, 2016.

\bibitem{yang2019xlnet}
Z.~Yang, Z.~Dai, Y.~Yang, J.~Carbonell, R.~R. Salakhutdinov, and Q.~V. Le,
  ``Xlnet: Generalized autoregressive pretraining for language understanding,''
  in \emph{Advances in neural information processing systems}, 2019, pp.
  5754--5764.

\bibitem{bahdanau2014neural}
D.~Bahdanau, K.~Cho, and Y.~Bengio, ``Neural machine translation by jointly
  learning to align and translate,'' \emph{arXiv preprint arXiv:1409.0473},
  2014.

\bibitem{schulman2017proximal}
J.~Schulman, F.~Wolski, P.~Dhariwal, A.~Radford, and O.~Klimov, ``Proximal
  policy optimization algorithms,'' \emph{arXiv preprint arXiv:1707.06347},
  2017.

\bibitem{mnih2016asynchronous}
V.~Mnih, A.~P. Badia, M.~Mirza, A.~Graves, T.~Lillicrap, T.~Harley, D.~Silver,
  and K.~Kavukcuoglu, ``Asynchronous methods for deep reinforcement learning,''
  in \emph{International conference on machine learning}, 2016, pp. 1928--1937.

\bibitem{konda2000actor}
V.~R. Konda and J.~N. Tsitsiklis, ``Actor-critic algorithms,'' in
  \emph{Advances in neural information processing systems}, 2000, pp.
  1008--1014.

\bibitem{schulman2015high}
J.~Schulman, P.~Moritz, S.~Levine, M.~Jordan, and P.~Abbeel, ``High-dimensional
  continuous control using generalized advantage estimation,'' \emph{arXiv
  preprint arXiv:1506.02438}, 2015.

\bibitem{bickel2009discriminative}
S.~Bickel, M.~Br{\"u}ckner, and T.~Scheffer, ``Discriminative learning under
  covariate shift.'' \emph{Journal of Machine Learning Research}, vol.~10,
  no.~9, 2009.

\bibitem{snoek2012practical}
J.~Snoek, H.~Larochelle, and R.~P. Adams, ``Practical bayesian optimization of
  machine learning algorithms,'' in \emph{Advances in neural information
  processing systems}, 2012, pp. 2951--2959.

\bibitem{adam}
D.~P. Kingma and J.~L. Ba, ``Adam: A method for stochastic optimization,'' in
  \emph{Proceedings of the International Conference on Learning
  Representations}, 2015.

\bibitem{harrison2015boston}
D.~Harrison and D.~Rubinfeld, ``Boston housing dataset,'' 2015.

\bibitem{coraddu2016machine}
A.~Coraddu, L.~Oneto, A.~Ghio, S.~Savio, D.~Anguita, and M.~Figari, ``Machine
  learning approaches for improving condition-based maintenance of naval
  propulsion plants,'' \emph{Proceedings of the Institution of Mechanical
  Engineers, Part M: Journal of Engineering for the Maritime Environment}, vol.
  230, no.~1, pp. 136--153, 2016.

\bibitem{bergstra2015hyperopt}
J.~Bergstra, B.~Komer, C.~Eliasmith, D.~Yamins, and D.~D. Cox, ``Hyperopt: a
  python library for model selection and hyperparameter optimization,''
  \emph{Computational Science \& Discovery}, vol.~8, no.~1, p. 014008, 2015.

\bibitem{you2018graph}
J.~You, B.~Liu, Z.~Ying, V.~Pande, and J.~Leskovec, ``Graph convolutional
  policy network for goal-directed molecular graph generation,'' in
  \emph{Advances in neural information processing systems}, 2018, pp.
  6410--6421.

\end{thebibliography}


\clearpage
\appendix
\section{Best Practices}
In order for better evaluation and reproduction of our research we address all the items in the checklist as mentioned in Lindauer and Hutter\cite{lindauer2019best}.
\begin{itemize}
    \item \textit{Code for the training pipeline used to evaluate the final
architectures} We used the search space of the architectures reported in NASBench-101 and thus the accuracy for all the architectures were precomputed. We have published code to train the policy network for ReMAADE algorithm, we also provide code for how different NAS algortihms were evaluated.

    \item \textit{Code for the search space} The publicly available NASBench-101 dataset search space was used.
    
    \item \textit{Hyperparameters used for the final evaluation pipeline,
as well as random seeds} The hyperparameters were left unchanged.

    \item \textit{For all NAS methods you compare, did you use exactly
the same NAS benchmark, including the same dataset,
search space, and code for training the architectures
and hyperparameters for that code?} Yes, as NASBench-101 was used for evaluation, the search space was fixed accordingly. All the different NAS methods we surveyed were evaluated against the same NASBench-101 dataset. 

    \item \textit{Did you control for confounding factors?}Yes, all the experiments across all NAS algorithms were on on the same NASBench-101 framework.
    
    \item \textit{Did you run ablation studies?} Yes, the results for the ablation studies have been outlined in the paper.
    
    \item \textit{Did you use the same evaluation protocol for the methods being compared?} Yes, the same evaluation protocol was used.
    
    \item \textit{Did you compare performance over time?} Yes, the performance was evaluated both in the short term and the medium term with an exploration budget of 150 architectures in the short term and an exploration budget of 3200 architectures in the medium-term..
    
    \item \textit{Did you compare to random search?} Yes.
    
    \item \textit{Did you perform multiple runs of your experiments
and report seeds?} Yes, we ran 500 trials of each experiment, with a different seed for each trial on NASBench-101. These results are completely reproducible.

    \item \textit{Did you use tabular or surrogate benchmarks for indepth evaluations } Yes, all our experiments were evaluated against the NASBench-101 dataset. 
    
    \item \textit{Did you report how you tuned hyperparameters, and
what time and resources this required?} We explored certain ranges of hyper-parameters get the best performance. These have been mentioned in the paper.

    \item \textit{Did you report the time for the entire end-to-end NAS
method?} Since all our experiments were run against the NASBench-101 framework for which we had pre-computed results, we were able to test the architectures generated by our network without training them from scratch. 

    \item \textit{Did you report all details of your experimental setup?}
    Yes, all the details for the experimental setup have been reported.
  
\end{itemize}

\end{document}